\documentclass[11pt,a4paper]{article}
\usepackage[hyperref]{emnlp2020}
\usepackage{times}
\usepackage{latexsym}

\aclfinalcopy 

\usepackage{microtype}
\usepackage{amsmath}
\usepackage{bm}
\usepackage{color}
\usepackage{multirow}
\usepackage{amsmath}
\usepackage{subfig}
\usepackage{enumitem}
\usepackage{amsfonts}
\usepackage{multirow}
\usepackage{makecell}
\usepackage{arydshln}
\usepackage{graphicx} 
\usepackage{graphics}
\usepackage{bbm}
\usepackage{balance}
\usepackage{algorithm}
\usepackage[noend]{algpseudocode}
\usepackage{booktabs}
\usepackage{textcomp}

\DeclareMathOperator*{\argmax}{arg\,max}

\title{Assessing the Bilingual Knowledge Learned by Neural Machine Translation Models}

\author{
Shilin He\thanks{~~Work done when interning at Tencent AI Lab.} \\ \normalsize The Chinese University of Hong Kong \\ \small \sf {slhe@cse.cuhk.edu.hk} \And
Xing Wang   \\ \normalsize Tencent AI Lab  \\ \small \sf {brightxwang@tencent.com} \AND
Shuming Shi \\  \normalsize Tencent AI Lab  \\ \small \sf {shumingshi@tencent.com} \And
Michael R. Lyu \\ \normalsize The Chinese University of Hong Kong   \\ \small \sf {lyu@cse.cuhk.edu.hk} \And
Zhaopeng Tu \\ \normalsize Tencent AI Lab  \\ \small \sf zptu@tencent.com
}

\begin{document}
\maketitle

\begin{abstract}
Machine translation (MT) systems translate text between different languages by automatically learning in-depth knowledge of bilingual lexicons, grammar and semantics from the training examples.
Although neural machine translation (NMT) has led the field of MT, we have a poor understanding on how and why it works. In this paper, we bridge the gap by assessing the bilingual knowledge learned by NMT models with {\em phrase table} -- an interpretable table of bilingual lexicons.
We extract the phrase table from the training examples that a NMT model correctly predicts.
Extensive experiments on widely-used datasets show that the phrase table is reasonable and consistent against language pairs and random seeds. Equipped with the interpretable phrase table, we find that NMT models learn patterns from simple to complex and distill essential bilingual knowledge from the training examples.
We also revisit some advances that potentially affect the learning of bilingual knowledge (e.g., back-translation), and report some interesting findings.
We believe this work opens a new angle to interpret NMT with statistic models, and provides empirical supports for recent advances in improving NMT models.

\end{abstract}

\section{Introduction}

Modern machine translation (MT) systems aim to produce fluent and adequate translations by automatically learning in-depth knowledge of bilingual lexicons, grammar and semantics from the training examples. 
Two technological advances solving this problem with statistical and neural techniques, statistical machine translation (SMT) and neural machine translation (NMT), have seen vast progress over the last two decades.
SMT models generate translations on the basis of several statistical models that {\em explicitly} represent the knowledge bases, such as translation model for bilingual lexicons, reordering and language models for grammar and semantics~\cite{koehn2009statistical}. Recently, NMT models have advanced the state-of-the-art by {\em implicitly} modeling the knowledge bases in a large neural network, which are trained jointly to maximize the translation performance~\cite{Bahdanau:2015:ICLR,gehring17:icml:2017,Vaswani:2017:NIPS}. 
Despite their power with a massive amount of parameters, we have limited understanding of how and why NMT models work, which poses great challenges for error analysis and model refinement.

In this work, we bridge the gap by assessing the knowledge bases learned by NMT models with the statistical models of SMT systems.
We believed (and in fact, provide some evidence to support the claim) that although using different forms (e.g., continuous vs. discrete) to represent the knowledge, NMT and SMT models are identical in modeling the essential knowledge. 
In the long-goal journey, we start with probing the bilingual knowledge with the translation model, also known as {\em phrase table}, which is one core component of SMT systems to represent the bilingual lexicons.
Bilingual knowledge is at the core of adequacy modeling, which is a major weakness of the NMT models~\cite{Tu_2016}.
Phrase table has proven its effectiveness for carrying useful bilingual knowledge, which can be seamlessly integrated to NMT models~\cite{Wang:2018:TASLP,lample2018phrase}. For instance, ~\newcite{lample2018phrase} have advanced the SOTA of unsupervised NMT by learning to align for phrase embeddings based on an external phrase table.

\begin{figure}
    \centering
    \subfloat[Output of an English $\Rightarrow$ German NMT model]{
    \includegraphics[width=0.48\textwidth]{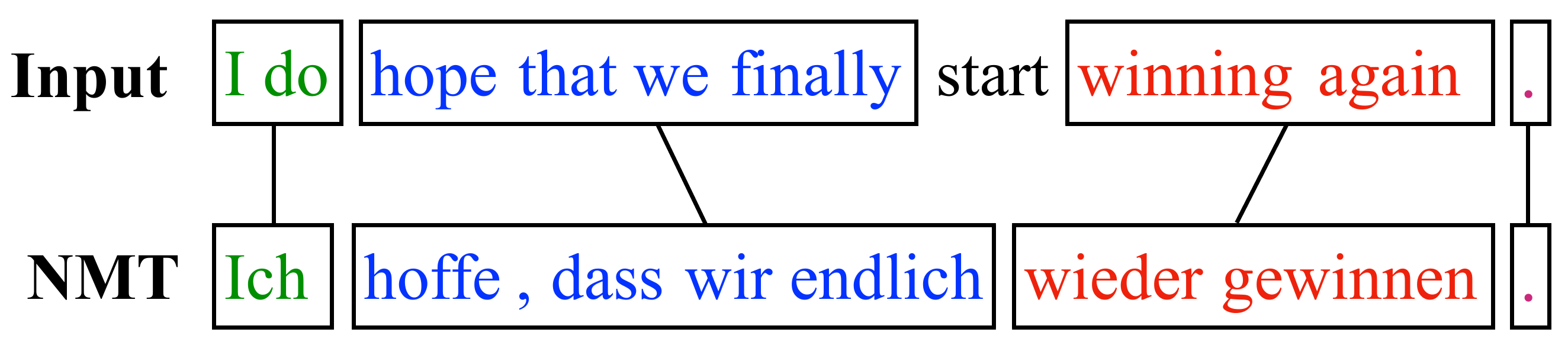}} \\
    \subfloat[Phrase table extracted from the NMT model]{
    \includegraphics[width=0.48\textwidth]{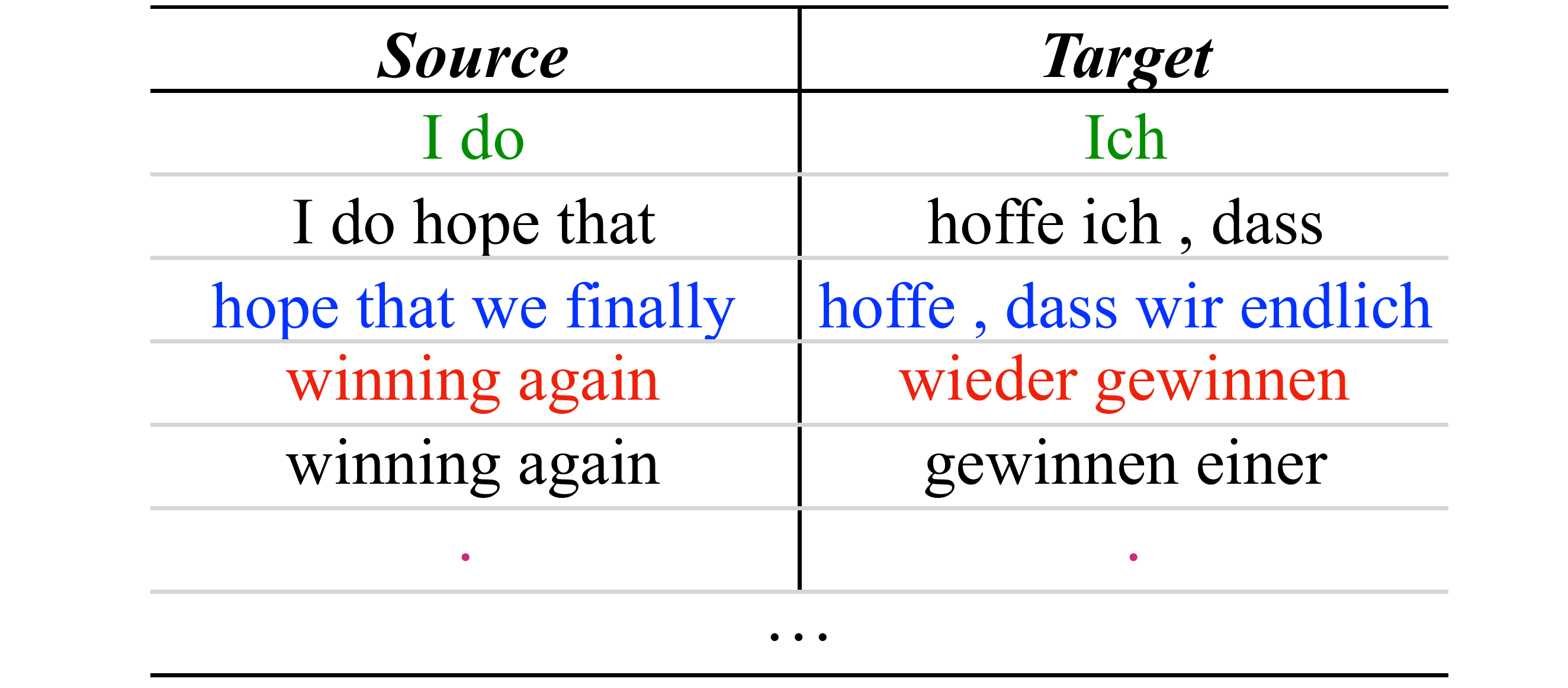}}
    \caption{The output of a NMT model (a) can be explained by the extracted phrase table (b). } 
    \label{fig:example}
\end{figure}

Specifically, we extract the phrase table from the predictions of NMT models, which is inspired by recent work on investigating the forgetting phenomenon of training examples in the image classification task~\cite{toneva:2018:empirical}. Intuitively, if a trained NMT model can successfully recover (part of) a training example, the NMT model is more likely to have learned the necessary bilingual lexicons for the recovery. 
Experimental results on three representative language pairs and random seeds show that the extracted phrase table correlates well with the NMT model performance, demonstrating that the phrase table can reasonably represent the bilingual knowledge learned by NMT models.
Figure~\ref{fig:example} shows an example, in which the phrase table extracted from a NMT model can well explain the generated output. 

With the interpretable phrase table in hand, we are able to better understand behaviors of NMT models in many aspects. We start with investigating the learning dynamics of bilingual knowledge. We find that NMT tend to first learn simple patterns and then complex patterns, and the catastrophic forgetting phenomenon occurs during the training.\footnote{We follow~\newcite{toneva:2018:empirical} to define ``forgetting event'' to have occurred when a training example transitions from being predicted correctly to incorrectly during training.}
We also reveal that one of the strengths of NMT models over SMT models lie in their ability to distill high-quality bilingual knowledge from the training data.

We then revisit some advances in improving NMT models, which potentially affect the learning of bilingual knowledge. Though we cannot claim causality, we have several observations:
\begin{itemize}
    \item {\em Model Capacity}: 
    We thought it likely that increasing model capacity learns more bilingual lexicons. This turned out to be false. Transformer-Big outperforms Transformer-Base by 1.3 BLEU points, while the extracted phrase tables are almost the same. We conjecture that the strengths of larger models lie in a better learning of more complex knowledge, such as composition rules to combine the bilingual lexicons.

    \item {\em Data Augmentation}: We investigate back-translation~\cite{sennrich:2016:acl} and forward-translation~\cite{zhang2016exploiting, he2019revisiting}, which introduce additionally synthetic parallel corpus. Both techniques improve performance not only by introducing new bilingual knowledge, but also with a better quality estimation of existing knowledge.
    
    \item {\em Domain Adaptation}: Fine-tune is a simple yet effective technique in domain adaptation, which learns to transfer out-domain knowledge to in-domain~\cite{luong2015stanford}. As expected, by adapting to the target-domain, the fine-tune approach learns more and better bilingual knowledge from the target-domain data.
\end{itemize}

\noindent The key contributions of this paper are:
\begin{itemize}
\item Our study demonstrates the reasonableness and effectiveness of assessing the NMT knowledge with statistic models, which opens up a new angle to interpret NMT models.

\item We report several interesting findings that can help humans better analyze and understand NMT models and some recent advances. 
\end{itemize}

\section{Related Work}

\paragraph{Evolution of MT Models.}
The MT task has a long history, in which the techniques have evolved from rule-based MT (RBMT)~\cite{hayes1985rule,sato1992example}, through SMT~\cite{brown:1993:CL,Och:2004:CL}, to NMT~\cite{Sutskever:2014:NIPS,Bahdanau:2015:ICLR}. RBMT methods require large sets of linguistic rules and extensive lexicons with morphological, syntactic, and semantic information, which are manually constructed by humans. Benefiting from the availability of large amounts of parallel data in 1990s, SMT approaches relieve the labor-intensive problem of RBMT by automatically learning the linguistic knowledge from bilingual corpora with statistic models.
More recently, NMT models have taken the field of MT by building a single network that can be trained on the corpora in an end-to-end manner. 
Several studies have shown that representations learned by NMT models contain a substantial amount of linguistic information on multiple levels: morphological~\cite{belinkov:2017:ACL}, syntactic~\cite{shi:2016:EMNLP}, and semantic~\cite{Hill:2017:MT}.

In the development circle of each generation, MT models are generally improved with techniques that are essential in the last generation. For example,~\newcite{Chiang:2005:ACL} and~\newcite{Liu:2006:ACL} relieved the nonfluent translation problem of SMT models by automatically learning syntactic rules from the parallel corpus, which are created manually by humans in RBMT systems.~\newcite{Tu_2016} alleviated the inadequate translation problem of NMT models by introducing the coverage mechanism, which is a standard concept in SMT to indicate how many source words have been translated.
Inspired by previous these studies, we hypothesize that MT models of different generations are possibly identical to model the essential knowledge. In this work, we propose to leverage the phrase table -- a basic module of SMT system, to assess the bilingual knowledge learned by NMT models.

\paragraph{Exploiting Phrase Table for NMT.}
Phrase table is an essential component of SMT systems, which records the correspondence between bilingual lexicons~\cite{koehn2009statistical}. 
Previous studies have incorporated phrase table as an external signal to guide the generation of NMT models~\cite{wang:2017:aaai,zhang2018prior,zhao2018phrase,guo2019non}. 
All these works show that the bilingual knowledge in phrase table can be identical to those in NMT models, and thereby can be seamlessly integrated to NMT models. Based on this observation, we employ the phrase table as an assessment tool of bilingual knowledge for NMT models. 

~\newcite{lample2018phrase} have advanced the SOTA of unsupervised NMT by evolving from learning alignment of word embeddings to learning to align for phrase embeddings based on an external phrase table, which is identical to the evolution of SMT from word-based model~\cite{brown:1993:CL} to phrase-based model~\cite{koehn:2003:NAACL}. This reconfirms our hypothesis that MT models of different generations are identical to model the essential knowledge, and thus share similar evolving trends.

\paragraph{Interpretability of NMT Models.}
The interpretability of NMT models has recently been approached mainly from two aspects~\cite{alvarez2017causal}: (1) {\em model interpretability}, which aims to understand the internal properties of NMT models, such as layer representations~\cite{shi:2016:EMNLP,belinkov:2017:ACL,Yang:2019:ACL,voita:2019:EMNLP} and attention~\cite{Voita:2019:ACL,Jain:2019:NAACL,Wiegreffe:2019:EMNLP,Li:2018:EMNLP}; and (2) {\em behavior interpretability}, which aims to explain particular behaviors of a NMT model, such as the input-output behavior~\cite{alvarez2017causal,ding2017visualizing,He:2019:EMNLP}. In this paper, we focus on the second thread from a complementary viewpoint -- assessing the bilingual knowledge learned by NMT models, which can provide explanations for model output, as shown in Figure~\ref{fig:example}.

\section{Assessing Bilingual Knowledge with Phrase Table}
\label{sec:bilingual knowledge}

In this section, we describe how to extract phrase table from the predictions of NMT models (Section~\ref{sec:extract_knowledge}), and verify our hypothesis by checking the correlation between the extracted phrase table and NMT performance (Section~\ref{section:evaluate_bilingual_knowledge}). 

\subsection{Methodology}
\label{sec:extract_knowledge}

There are many possible ways to implement the general idea of extracting phrase table from the predictions of NMT models. The aim of this paper is not to explore this whole space but simply to show that one fairly straightforward implementation works well and the proposed framework is reasonable. We leave the exploitation of more advanced forms of statistic models on bilingual knowledge such as syntax rules~\cite{Liu:2006:ACL} and discontinuous phrases~\cite{Galley:2010:NAACL} for future work.

We follow the standard pipeline in SMT to construct the phrase table with a two-phase approach.
The first phase, which is the focus of this paper, is {\em phrase extraction} where the bilingual phrase pairs are extracted from a word-aligned parallel data. Secondly, each phrase pair is assigned with some scores, which are estimated based on the occurrences of these phrases or their words on the same word-aligned training data.
The key challenge lies in how to incorporate the prior of NMT predictions into the SMT pipeline. In this study, we model the NMT priors as a mask sequence, which is integrated into the standard SMT pipeline as a constraint, as listed in Algorithm~\ref{alg:phrase-table}. 

\begin{algorithm}[t]
\small
\caption{Constructing Phrase Table}
\label{alg:phrase-table}
\begin{algorithmic}[1]
\Statex \textbf{Input}: training example ($\bf x$, $\bf y$), alignment $\bf a$, mask $\bf m$
\Statex \textbf{Output}: phrase set $\mathcal{R}$

\Procedure {PhraseTable}{}
\State \textsc{Extraction}
\State \textsc{Estimation}
\EndProcedure

\Procedure {Extraction}{}
\State $\widehat{\mathcal{R}} \leftarrow$ extract candidates from \{($\bf x$, $\bf y$), $\bf a$\}
\For {each $r \in \widehat{\mathcal{R}}$}   \Comment{{\em priors of NMT predictions}}
    \If {$r$ is consistent with $\bf m$}  
        \State $\mathcal{R}$.append($r$)
    \EndIf
\EndFor
\EndProcedure

\Procedure {Estimation}{}
\State standard procedure
\EndProcedure
\end{algorithmic}
\end{algorithm}

\paragraph{Building {\em Masked} Word-Aligned Parallel Data.}
Inspired by~\newcite{toneva:2018:empirical}, we define ``{\em memorized phrase pair}'' to be extracted from the associated (partial) training example, which is predicted correctly by the NMT model.
To this end, we first decompose the sequence generation of NMT into a series of classification tasks. Given a training example $(\textbf{x} = \{x_1, \dots, x_I\}, \textbf{y} = \{y_1, \dots, y_J\})$ and a model $M$, we use the model $M$ to force-decode $\textbf{x}$ to $\textbf{y}$, and check whether each $y_j$ is correctly predicted by $M$:
\[m_j= 
\begin{cases}
    1, & \text{if } y_j = \argmax_y \mathcal{N}_j[y] \\
    0,              & \text{otherwise}
\end{cases}
\]
where $\mathcal{N}_j$ is the probability distribution of model prediction at step $j$. A token $y_j$ is predicted correctly if it is assigned the highest probability by the model (``$y_j = \argmax_y \mathcal{N}_j[y]$''). 

Intuitively, a token $y_j$ with mask $m_j=0$ denotes that this token is not correctly predicted by the model. Accordingly, any phrase pairs that contain the token $y_j$ should not be extracted from the training example $(\textbf{x}, \textbf{y})$, since these phrase pairs are not fully learned by the NMT model. A lightweight implementation is to replace these tokens with a special symbol ``\$MASK\$'', and run the standard phrase extraction phase as in the SMT pipeline. Then we remove all the phrase pairs that contain the symbol ``\$MASK\$'' (lines 6-8 in Algorithm~\ref{alg:phrase-table}), and feed the pruned phrase pairs to the second phase of parameter estimation.

\begin{figure*}[t]
    \centering
    \subfloat[En$\Rightarrow$De]{
    \includegraphics[width=0.3\textwidth]{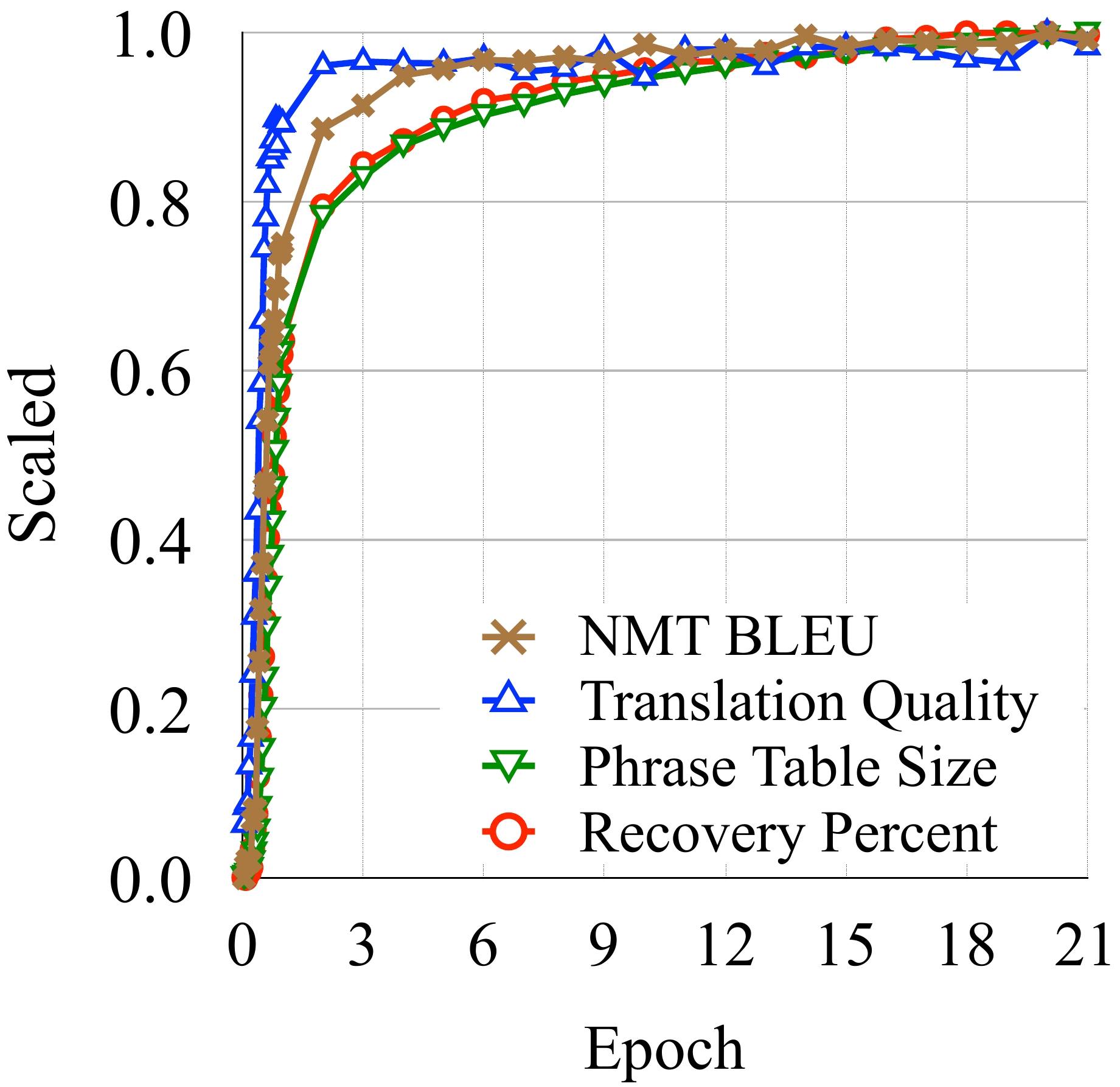} \label{fig:phrase_en_de}
    } \hfill
    \subfloat[En$\Rightarrow$Ja]{
    \includegraphics[width=0.3\textwidth]{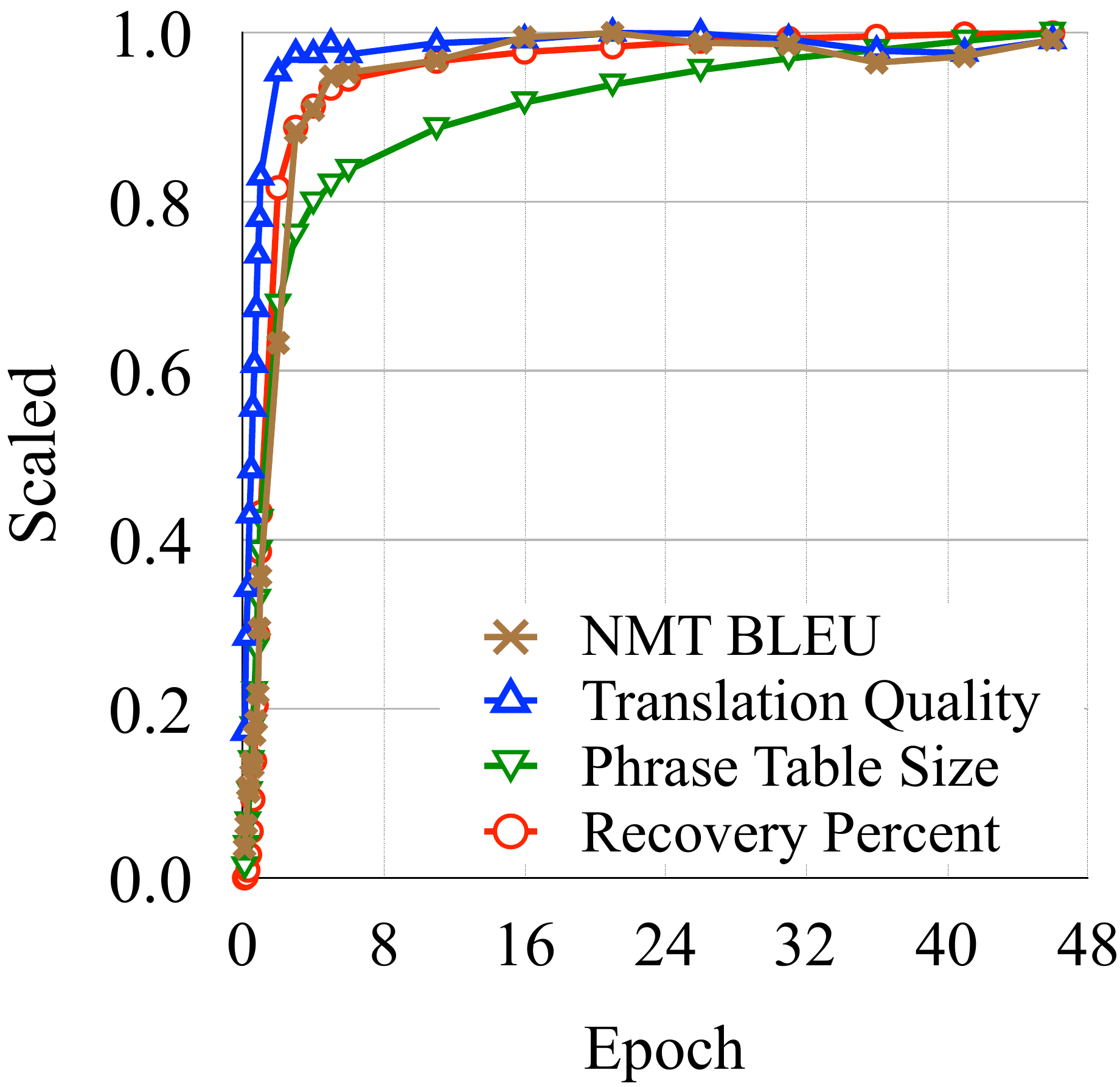} \label{fig:phrase_en_ja} 
    }\hfill
    \subfloat[En$\Rightarrow$De with Different Seeds]{
    \includegraphics[width=0.28\textwidth]{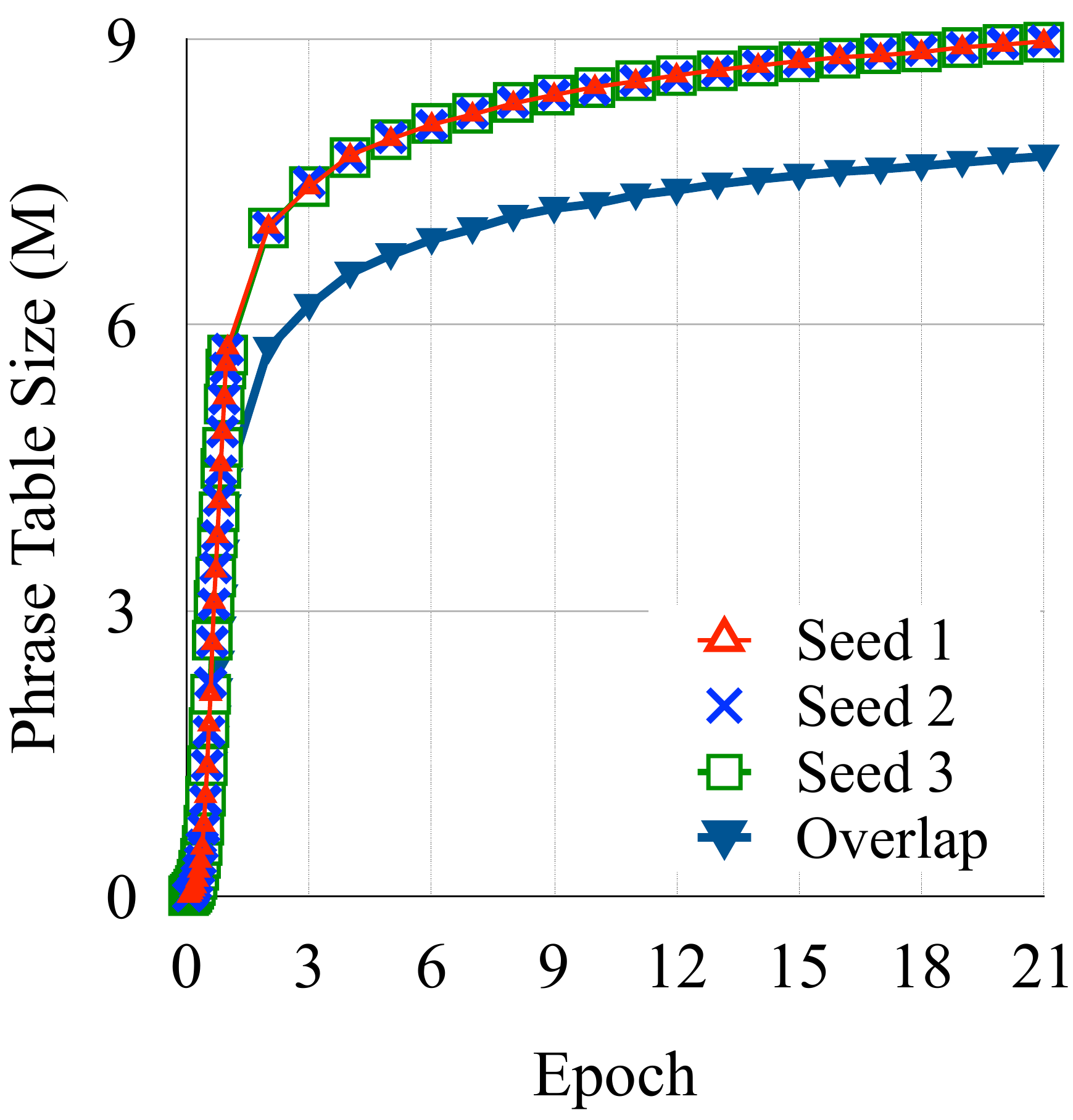} \label{fig:phrase_seeds}
}
\caption{Evaluating the correlation between quality metrics of phrase table and NMT performance (``NMT BLEU'') on (a) En$\Rightarrow$De and (b) En$\Rightarrow$Ja datasets. All metrics are scaled by the corresponding best score to fit in the figure. We also report results on phrase table size for NMT models with different random seeds (c). }
\label{fig:phrase_table}
\end{figure*}

\subsection{Experimental Setup}
\label{section:setup}

\paragraph{Data and Models}
We conduct experiments on both the widely-used WMT2014 English$\Rightarrow$German (En$\Rightarrow$De) and the syntactically-distant WAT2017 English$\Rightarrow$Japanese (En$\Rightarrow$Ja)~\cite{neubig:2015:wat} datasets. We use 4-gram NIST BLEU score~\cite{papineni2002bleu} as the evaluation metric.

{For SMT experiments, we follow the standard SMT pipeline and the setting of Edinburgh’s phrase-based system in WMT-2014~\cite{durrani2014edinburgh} with as few human heuristics as possible.} 
We use Moses~\cite{koehn2007moses} with default system setting and the toolkit Fast\_Align~\cite{dyer:2013:naacl} for building word-aligned corpus, which is fast and automatic.
Following~\newcite{johnson2007improving} to reduce redundancy, we further remove phrase pairs that occur only once in the training data.

For NMT experiments, We use the toolkit Fairseq~\cite{ott:2019:naacl} to implement NMT models~\cite{Vaswani:2017:NIPS}. We train the NMT models for 100,000 steps and save the checkpoint models at each epoch. In the first epoch, we save the model per 200 steps and extract phrase tables from training examples that have seen on so far only.

\paragraph{Evaluation Metrics} To verify our claim in this paper, we propose several metrics to quantitatively evaluate quality of the phrase table. If the metrics correlate well with NMT performance, then the phrase table is a reasonable assessment to represent the bilingual knowledge learned by NMT models.
The metrics are as follows:

\vspace{3pt}
\noindent \textit{Phrase Table Size:} As a straightforward metric, the size measures the number of distinct phrase pairs in a phrase table. A larger phrase table size indicates more abundant bilingual knowledge.

\vspace{3pt}
\noindent \textit{Recovery Percent:} The phrase table size might be less accurate due to duplicate counting of compositions of existing phrase pairs. Accordingly, we propose another metric, recovery percent, to measure the distinct knowledge on data reconstruction. In detail, we use the phrase table to force decode the target sentence to recover as many target tokens as possible, and the ratio is denoted as the recovery percentage. A higher recovery percent indicates more distinct knowledge since more data can be reconstructed based on the phrase table.

\vspace{3pt}
\noindent \textit{Translation Quality:} Finally, we directly evaluate the essential knowledge in the phrase table for the ultimate translation. Specifically, we train a SMT model with the extracted phrase table by the off-the-shelf Moses toolkit, and evaluate its BLEU score on the test set. For fair comparison, we keep other SMT components unchanged and only alter the phrase table, therefore the relative SMT BLEU values is our focus of interest.

\subsection{Evaluating the Phrase Table}
\label{section:evaluate_bilingual_knowledge}

\paragraph{The extracted phrase table correlates well with the NMT performance.} Figure~\ref{fig:phrase_en_de} illustrates the results of the above metrics on the English$\Rightarrow$German dataset.
As seen, all three metrics are highly in line with the NMT performance (``NMT BLEU'') during the entire learning process. 
The Pearson correlations between NMT BLEU scores and phrase table size, recovery percent, and the translation quality are 0.975, 0.987, and 0.956, respectively, demonstrating very high correlations between the phrase table and NMT performance. This confirms our claim that phrase table is a reasonable assessment to represent the bilingual knowledge learned by NMT models.

\paragraph{The conclusion is robust across language pairs and random seeds.} 
We also validate our approach on the English$\Rightarrow$Japanese dataset, as shown in Figure~\ref{fig:phrase_en_ja}. 
The Pearson correlations are respectively 0.988, 0.990, and 0.908, demonstrating the universality of our conclusions .
To avoid the potential bias, we vary the initialization seed and analyze whether the extracted phrase table is robust. Figure~\ref{fig:phrase_seeds} depicts the results. The phrase table size increases similarly in different seeds. Additionally, at each epoch, more than 85\% phrase pairs are same among three seeds  (``Overlap''), which shows that its robustness against random seeds. 

Given the general applicability of the phrase table, we use the English$\Rightarrow$German translation as our test bed for further analyses. We will interchangeably use the terms ``phrase table'' and ``bilingual knowledge'' in the following sections.

\section{Learning of Bilingual Knowledge}
\label{sec:analyze}
With the interpretable phrase table in hand, we attempt to understand how NMT models learn the bilingual knowledge from two perspectives:
\begin{itemize}
    \item How do NMT models learn the bilingual knowledge during training? (Section~\ref{sec:learning_dynamics})
    \item Does the trained NMT model sufficiently explore the bilingual knowledge embedded in the training examples? (Section~\ref{sec:knowledge})
\end{itemize}

\subsection{Learning Dynamics}
\label{sec:learning_dynamics}

In this section, we investigate the evolvement of bilingual knowledge during the training. To this end, we first categorize the phrase pair into different complexity levels using several metrics that are widely used in the SMT research:

\vspace{5pt}
\noindent {\em Phrase Length:} A longer phrase is usually of more complexity~\cite{lu2010automatic}. We categorize the phrase length into three types with increasing complexity: {\em short} (1-3) $<$ {\em middle} (3-5) $<$ {\em long} (5-7).

\vspace{5pt}    
\noindent {\em Reordering Type:} 
This metric measures the order of two phrases with lexicalized reordering~\cite{tillmann:2004:naacl}, and disordered phrases are often hard to translate~\cite{koehn2009statistical}. We have three types with increasing complexity: {\em monotone} $<$ {\em swap} $<$ {\em discontinuous}.

\vspace{5pt}    
\noindent {\em Word Fertility:} Word fertility measures the alignment relations between the words inside the phrase pair. Words with a complex fertility might indicate inherent translation difficulty~\cite{brown1990statistical}. We have three fertility types with increasing difficulty: {\em 1-1 align} $<$ {\em M-1 align} $<$ {\em 1-M align}.

\begin{figure*}[t]
    \centering
    \subfloat[Phrase Length]{
    \includegraphics[width=0.3\textwidth]{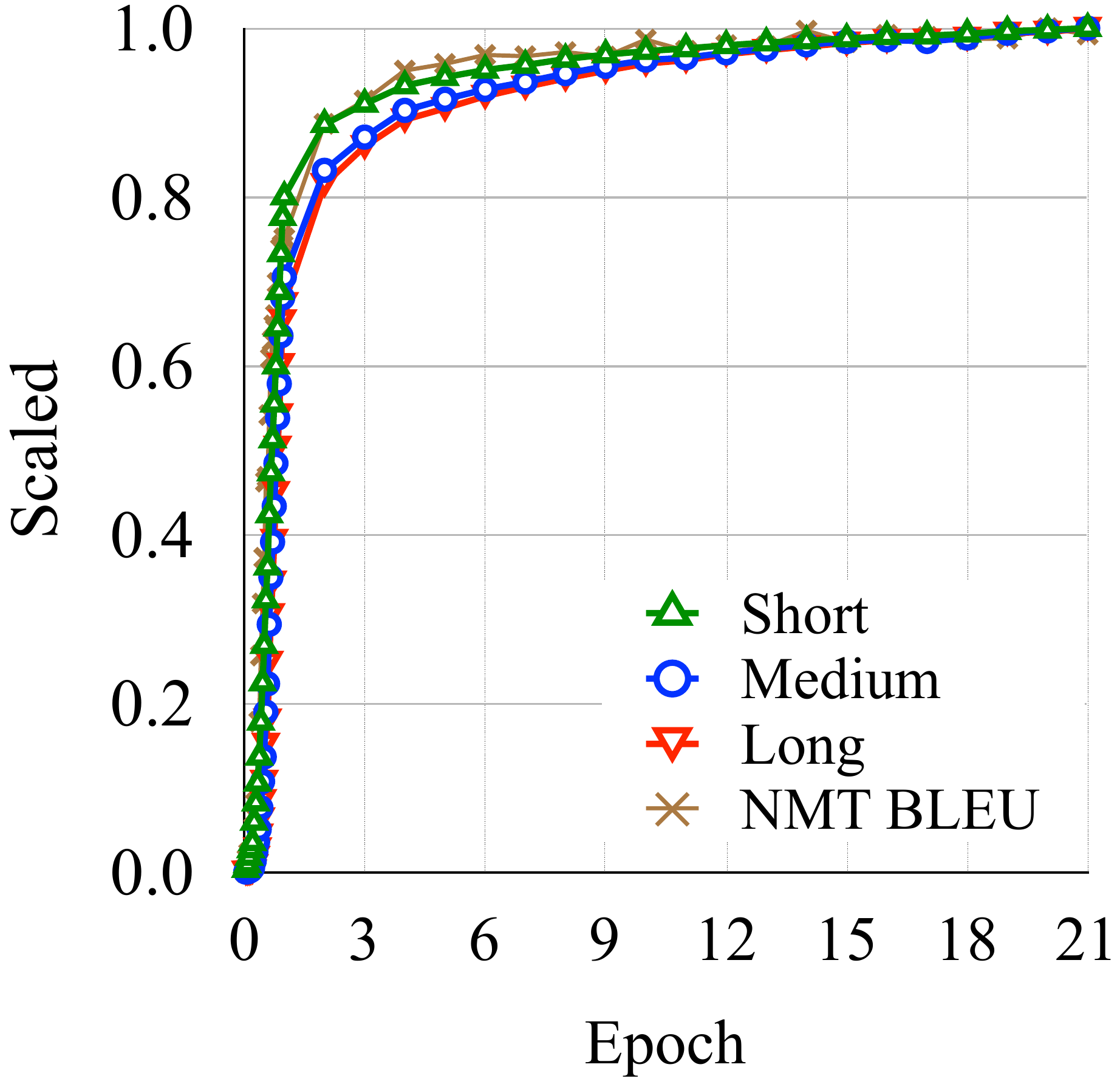} \label{fig:phrase_length}
    } \hfill
    \subfloat[Reordering Type]{
    \includegraphics[width=0.3\textwidth]{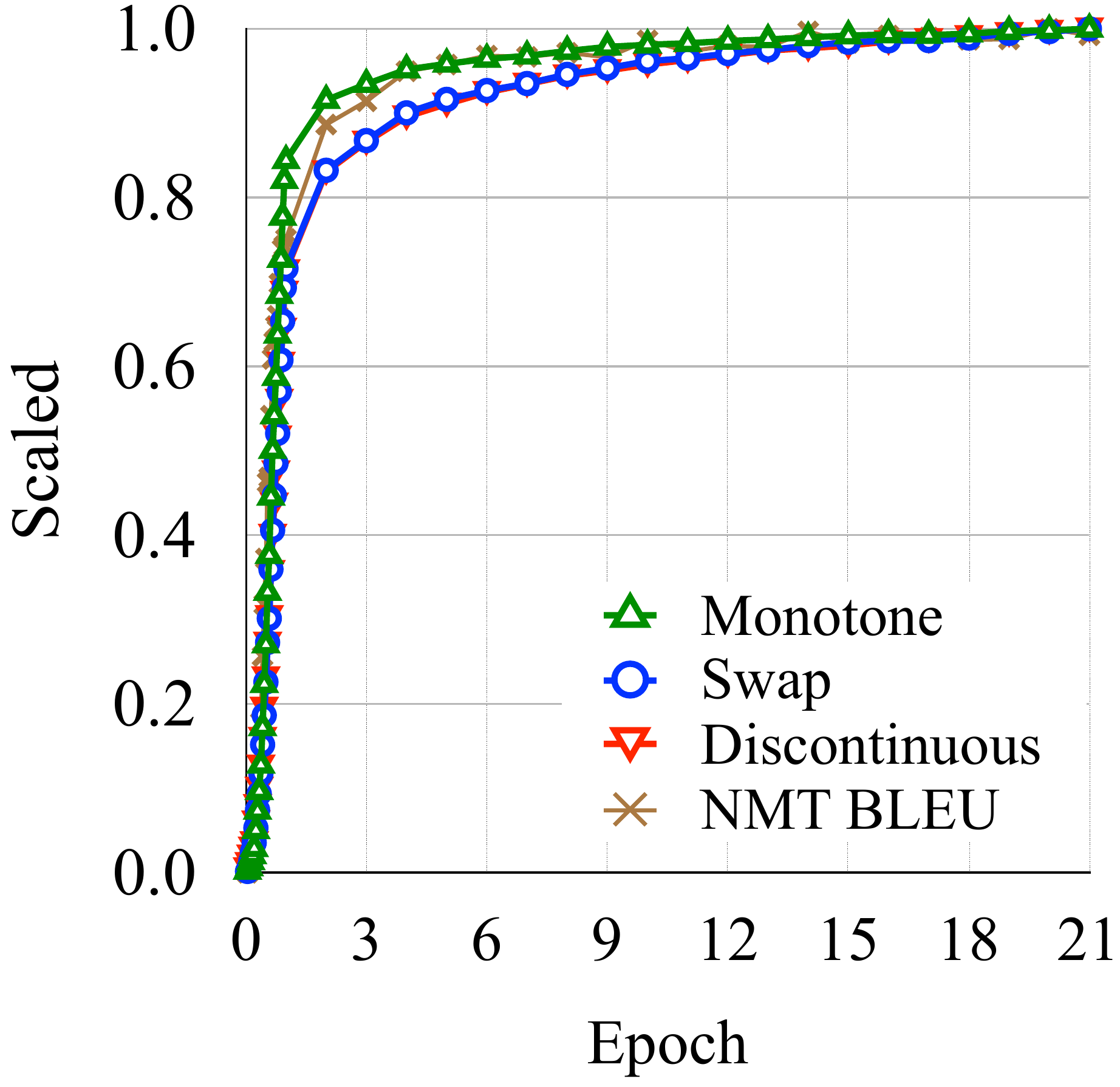} \label{fig:phrase_reorder}
}  \hfill
    \subfloat[Word Fertility]{
    \includegraphics[width=0.3\textwidth]{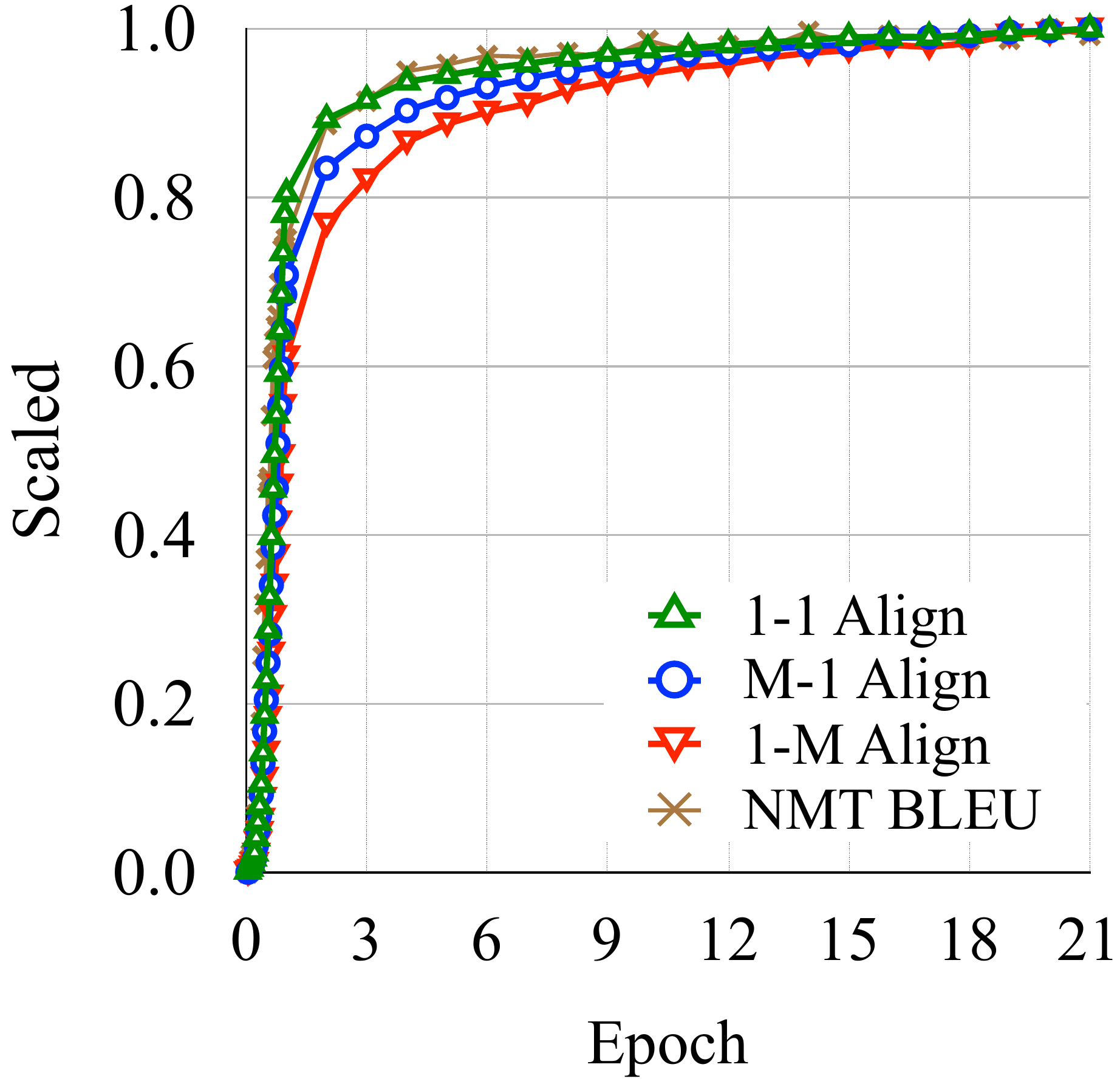} \label{fig:phrase_fertility}
}
\caption{Learning dynamics of bilingual knowledge according to different complexity metrics.}
\label{fig:learning-dynamics}
\end{figure*}

For each metric, we normalize the value by the maximum phrase pair size in each category.

\paragraph{NMT models tend to learn simple patterns first and complex patterns later.} As shown in Figure~\ref{fig:phrase_length}, NMT models learn short phrases faster than medium phrases and long phrases, embodied by a fastest convergence and a highest slope among three categories in the first epoch. As the learning continues, medium and long phrases start to converge to a relative stable state slowly. Besides, NMT BLEU scores show a very similar increasing trend as the short phrase, demonstrating a high correlation (Pearson correlation: 0.992) between the NMT performance and short phrases.

We can observe similar findings on the phrase reordering type (Figure~\ref{fig:phrase_reorder}) and word fertility (Figure~\ref{fig:phrase_fertility}). Simple patterns like monotone and 1-1 aligned phrase can be quickly learned by NMT models, while complex patterns are learned in a slower manner.
This is in line with the findings of~\newcite{Rohaman:2019:ICML}: deep networks will first learn low-complexity functional components, before absorbing high-complexity features.
These results also indicate that NMT models might by nature has the learning ability similar to the \textit{curriculum learning}~\cite{bengio2009curriculum,kocmi2017curriculum} without any explicit curriculum.

\paragraph{Forgetting dynamics occur in the learning of bilingual knowledge.}
As shown in Figures~\ref{fig:phrase_table} and~\ref{fig:learning-dynamics}, the size of learned phrase table is monotonically increasing as the learning processes. One question naturally arises: {\em are the phrase pairs never forgotten once learnt?}

\begin{figure}[t]
    \centering
    \includegraphics[width=0.30\textwidth]{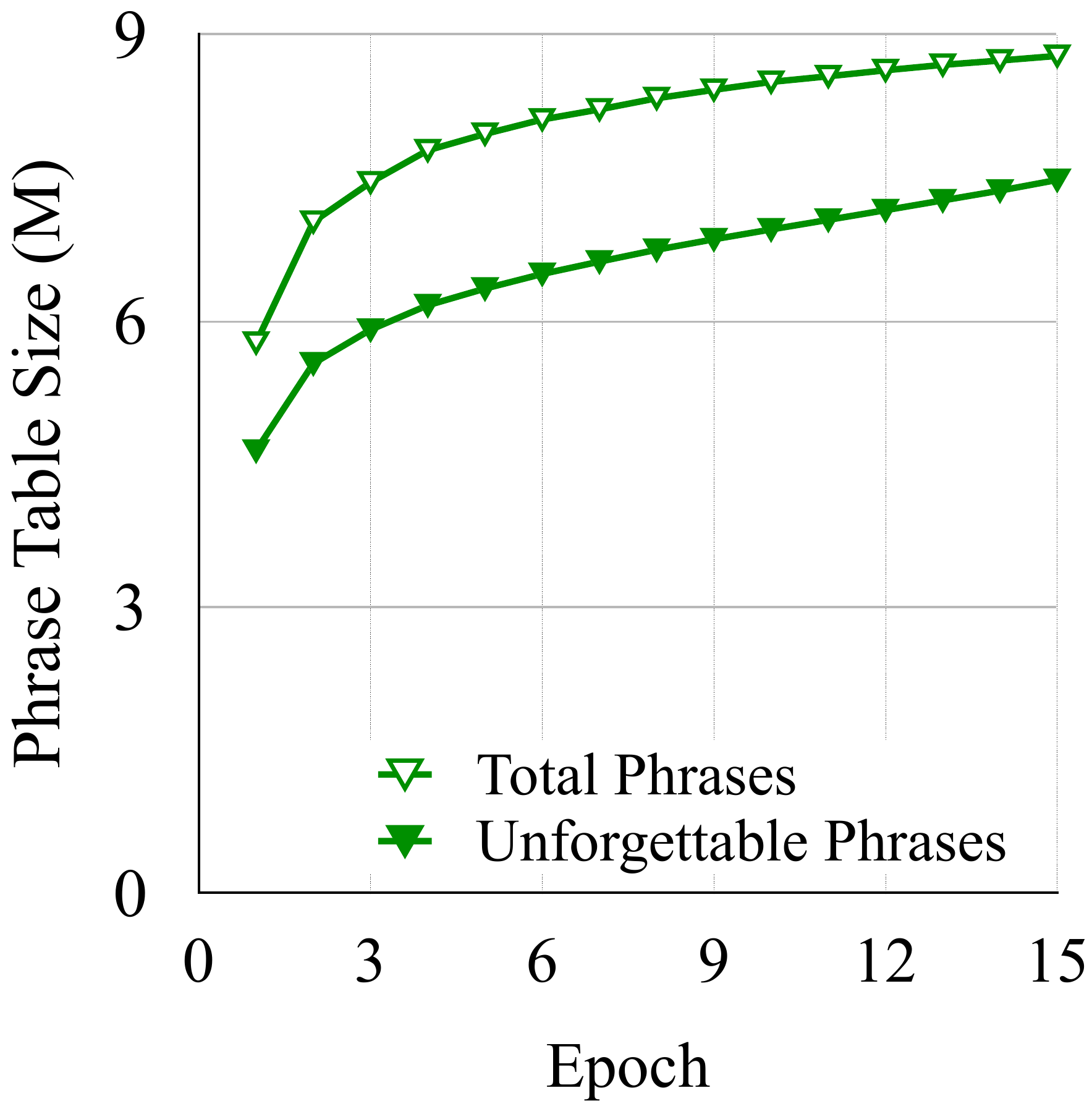}
\caption{Illustration of unforgettable phrases.}
\label{fig:phrase_forget}
\end{figure}

Figure~\ref{fig:phrase_forget} shows the result. Note that we only plot the first 15 epochs to ensure that the phrases are never forgotten for at least 6 epochs. 
Around 80\% of learned phrase table is unforgettable phrases (always learned phrase pairs), while the rest phrase pairs are forgotten. This is consistent with the findings of~\newcite{toneva:2018:empirical} on the image classification tasks.

\begin{table*}[t]
\begin{center}
 \begin{tabular}{c|r|r|r|r|r|r}
  \multirow{2}{*}{\bf Phrase Table}  &  \multicolumn{2}{c|}{\bf Shared} &   \multicolumn{2}{c|}{\bf Non-Shared}  &  \multicolumn{2}{c}{\bf All}\\
  \cline{2-7}
        &  \em Size    & \em BLEU  &   \em Size   & \em BLEU  &   \em Size    & \em BLEU\\
  \hline
  \hline
  Full      &   9.0M   &   17.32   &   8.5M   &   4.50    &  { 17.5M }   &    17.91\\
  \hline
  NMT       &   9.0M   &  17.90  &   0M   &   0   &   {9.0M }  & 17.90\\
 \end{tabular} 
\end{center}
\caption{Comparison of the phrase table extracted from the full training data (``Full'') and NMT models (``NMT''). 
``All'' denotes the whole phrase table, ``Shared'' denotes the intersection of two tables, and ``Non-shared'' denotes the complement. Note that the probabilities of ``Shared'' phrases are different for the two tables.}
\label{table:knowledge-distillation}
\end{table*}

\subsection{Learned Bilingual Knowledge}
\label{sec:knowledge}

In this experiment, we evaluate whether NMT models have sufficiently explore the bilingual knowledge in the training examples, by comparing the phrase tables extracted from NMT predictions and from the raw training data. We use the latter to represent the full bilingual knowledge embedded in the training examples.

As shown in Table~\ref{table:knowledge-distillation}, the bilingual knowledge learned by NMT model (``NMT'') shows comparable translation quality with the full-data knowledge (``Full'') (17.90 vs. 17.91), but with only {\em a half of } phrases (9.0M vs. 17.5M).\footnote{When considering the filtered one-shot phrases, NMT phrase only takes 22.8\% of the full table (76M vs. 335M).}
In addition, NMT provides a better probability estimation for the distilled phrases (``Shared'', 17.90 vs. 17.32). In the ``Non-Shared'' table, 78.2\% of the phrase pairs share the same source phrase with the ``Shared'' table, of which 83.2\% have a lower translation probability. The results empirically confirm our hypothesis that {\em NMT models distill the bilingual knowledge by discarding those low-quality phrase pairs}.

\section{Revisiting Recent Advances}
\label{sec:application}

In this section, we revisit recent advances that potentially affect the learning of bilingual knowledge. Specifically, we investigate three types of techniques: (1) {\em model capacity} that indicates how complicated patterns a model can express (Section~\ref{sec:capacity}); (2) {\em data augmentation} that introduces additional knowledge with external data (Section~\ref{sec:augmentation}); and (3) {\em domain adaptation} that transfers knowledge across different domains (Section~\ref{sec:adaptation}).

\subsection{Model Capacity }
\label{sec:capacity}

\begin{table}[t]
  \centering
  \begin{tabular}{c||r|c||r|c}
  \multirow{2}{*}{\bf Model}  &
  \multicolumn{2}{c||}{\bf NMT} &
  \multicolumn{2}{c}{\bf Phrase Table} \\
  \cline{2-5}
     & \em {\#Para}  & \em  BLEU & \em Size & \em BLEU\\
  \hline
  \textsc{Small} & 38M   & 25.45  & 7.7M & 17.35\\  
  \textsc{Base}  & 98M   & 27.11 &  9.0M & 17.90 \\
  \textsc{Big}   & 284M  & 28.40 &  9.2M & 17.89 \\
    \end{tabular}
  \caption{Results of NMT models of different capacities.}
  \label{tab:capacity}
\end{table}

\begin{table}[t]
\begin{center}
 \begin{tabular}{c|r|r||r|r}
  \multirow{2}{*}{\bf Model}  &  \multicolumn{2}{c||}{\bf Shared} &   \multicolumn{2}{c}{\bf Non-Shared} \\
  \cline{2-5}
        &  \em Size    & \em BLEU  &   \em Size   & \em BLEU  \\
  \hline
  \hline
  \textsc{Small}  & 7.0M  &  17.53 & 0.7M & 2.37 \\
  \textsc{Base}   & 7.0M  &  17.49 & 2.0M & 3.57\\
  \textsc{Big}    & 7.0M  &  17.29 & 2.2M & 3.47 \\
 \end{tabular} 
\end{center}
\caption{Comparison of phrases from three capacities.}
\label{tab:capacity-shared}
\end{table}

We vary the layer dimensionality of Transformer, and obtain three model variants: \textsc{Small} (256), \textsc{Base} (512), and \textsc{Big} (1024). As listed in Table~\ref{tab:capacity}, increasing model capacity consistently improves translation performance. However, the extracted phrase table is only marginally increased.

We compare the phrase tables learned by different models, as shown in Table~\ref{tab:capacity-shared}. The phrases shared by all models take the overwhelming majority, which add most value to the translation performance. We conjecture that enlarging capacity improves NMT performance by better exploiting complex patterns beyond bilingual lexicons. This also confirms our intuition that bilingual lexicons can be a crucial early step in assessing the knowledge in NMT models.

\subsection{Data Augmentation}
\label{sec:augmentation}

\begin{table}[t]
  \centering
  \begin{tabular}{l||r|c||r|c}
  \multirow{2}{*}{\bf Model}  &
  \multicolumn{2}{c||}{\bf NMT} &
  \multicolumn{2}{c}{\bf Phrase Table} \\
  \cline{2-5}
    \bf    & \em {\#Para}  & \em  BLEU & \em Size & \em BLEU\\
    \hline 
    \hline
    \textsc{Base}  & 98M   & 27.11 &  9.0M & 17.90 \\
    \hline
     ~~~ + BT         & 98M   & 29.75 & 20.9M & 19.26 \\
     ~~~ + FT         & 98M   & 28.43 & 28.0M & 19.33 \\
    \end{tabular}
  \caption{Results of back-translation (``BT'') and forward-translation (``FT'').}
  \label{tab:augmentation}
\end{table}

\begin{table}[t]
\begin{center}
 \begin{tabular}{c|r|r||r|r}
  \multirow{2}{*}{\bf Model}  &  \multicolumn{2}{c|}{\bf Shared} &   \multicolumn{2}{c}{\bf Non-Shared} \\
  \cline{2-5}
        & \em Size    &  \em BLEU  & \em  Size   & \em BLEU   \\
  \hline \hline
  \textsc{Base}      & 8.3M  & 17.67  & 0.7M    &  1.78   \\
  ~~~ + BT           & 8.3M  & 18.61  & 12.6M   &  10.45    \\
  \hline \hline
  \textsc{Base}      & 8.4M  & 17.83    & 0.5M  & 1.21 \\
  ~~~ + FT           & 8.4M  & 18.30    & 19.6M & 11.25 \\

 \end{tabular} 
\end{center}
\caption{Comparison of phrases learned by BT and FT.}
\label{table:back_transaltion_shared}
\end{table}

\begin{figure}[t]
    \centering
    \subfloat[Length]{
    \includegraphics[height=0.22\textwidth]{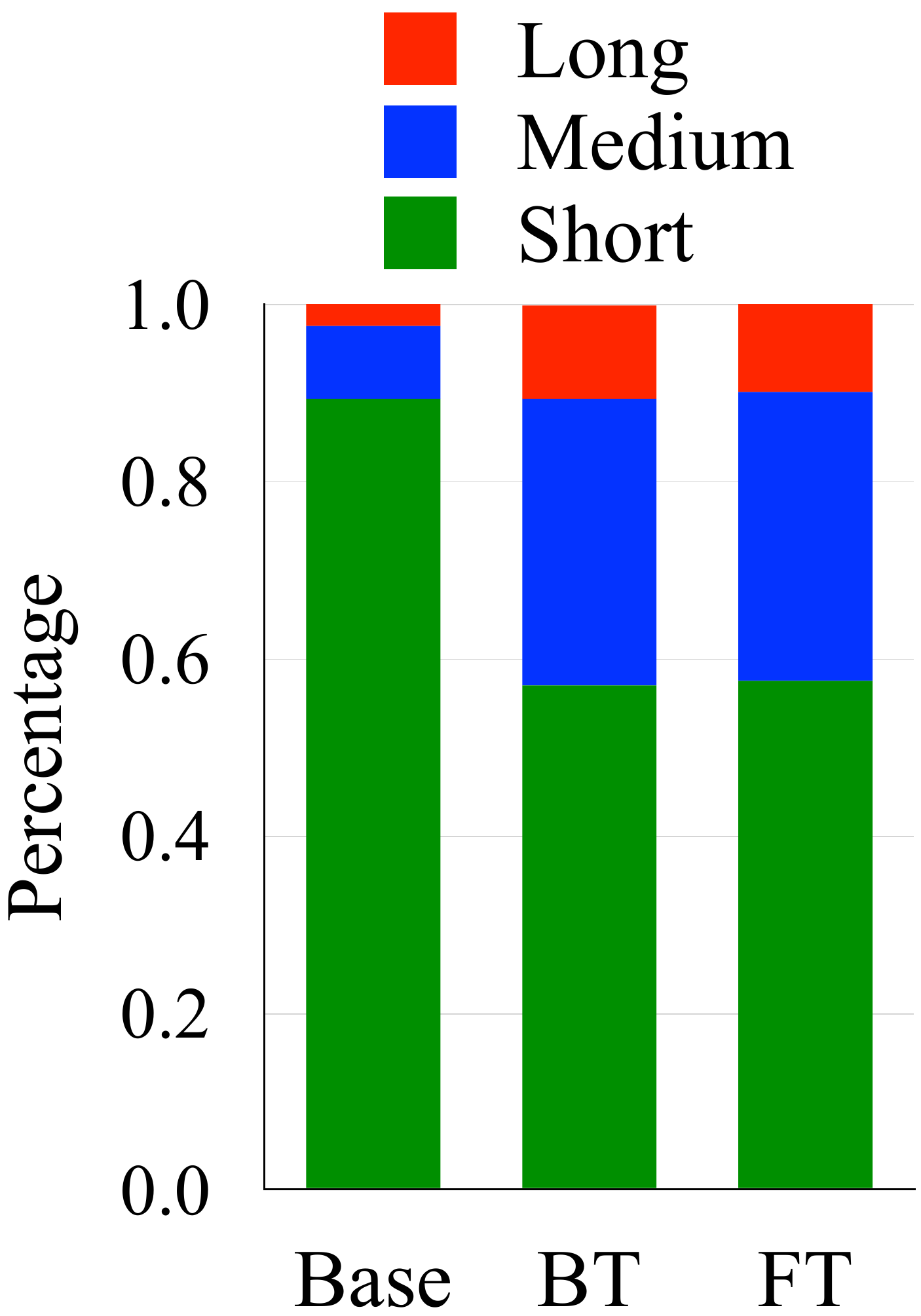} } \hfill
    \subfloat[Reordering]{
    \includegraphics[height=0.22\textwidth]{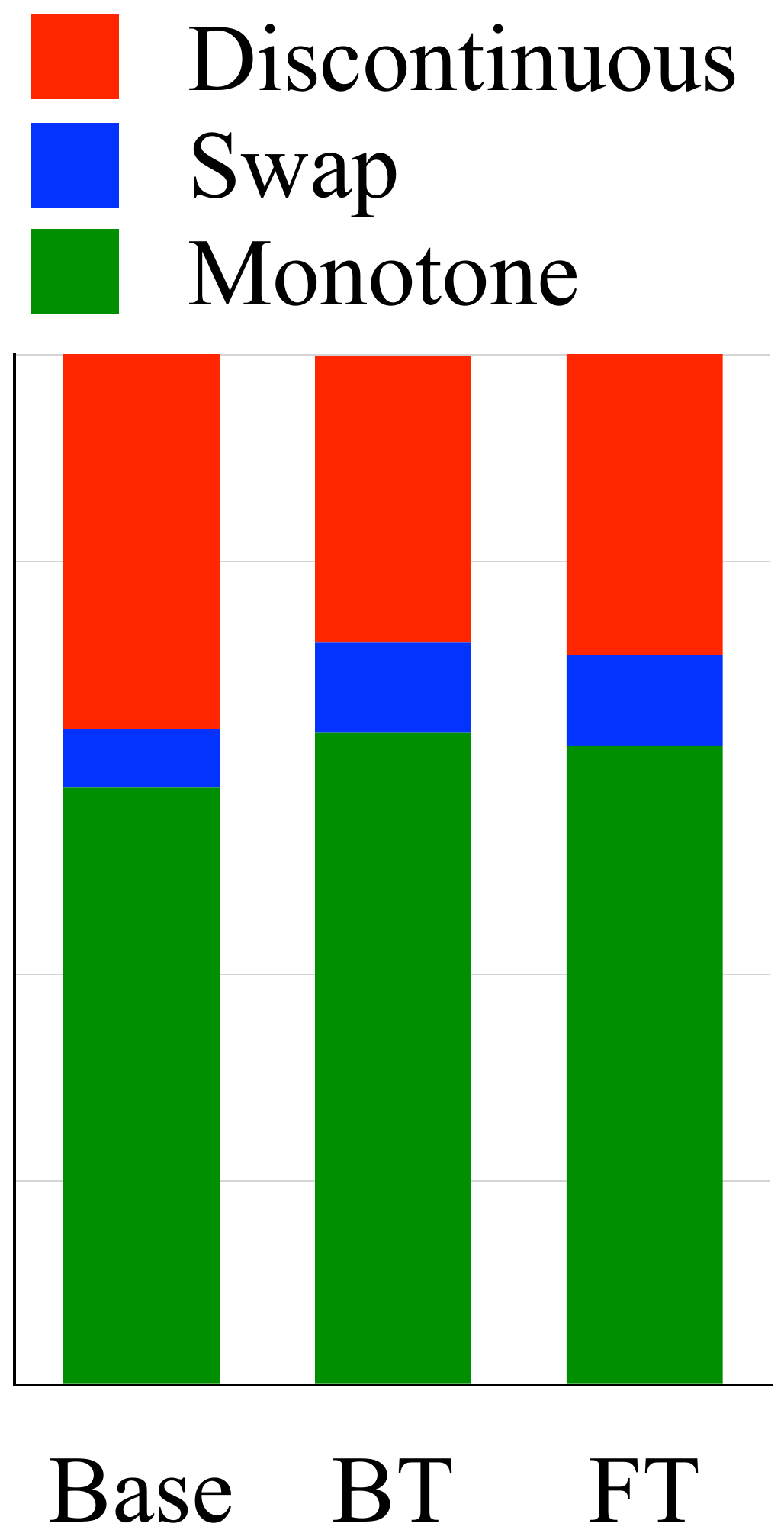} } \hfill
    \subfloat[Fertility]{
    \includegraphics[height=0.22\textwidth]{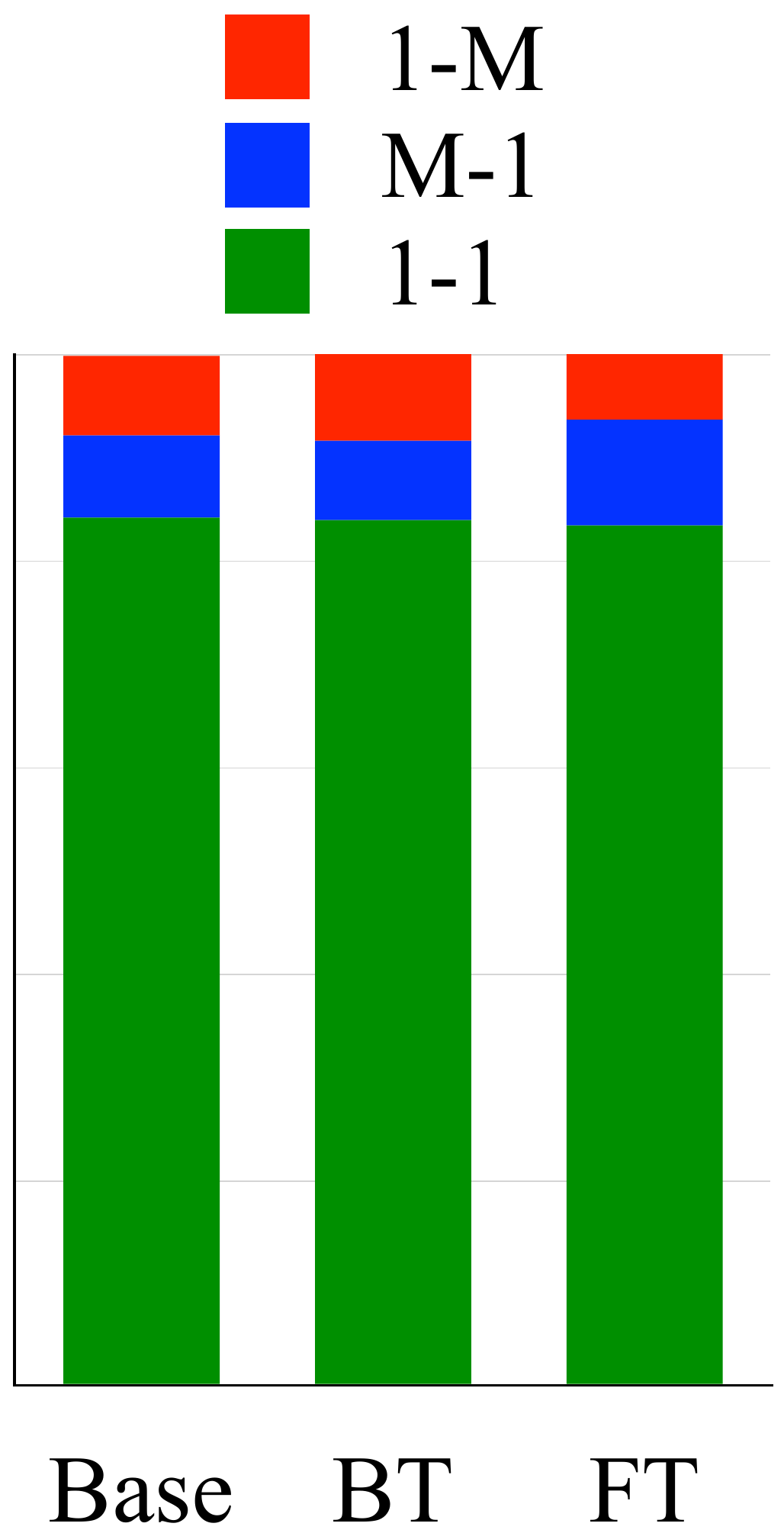} }
\caption{Characteristic distributions of phrases newly introduced by BT and FT. For a better illustration, we also give the distribution of the baseline (``Base'').}
\label{fig:back_translation_shared}
\end{figure}

In this experiment, we investigate two representative data augmentation approaches, back-translation~\cite{sennrich:2016:acl} and forward-translation~\cite{zhang2016exploiting}, which differ at exploiting target or source-side monolingual data. We select a same-size (around 4.5M) English and German monolingual dataset from the WMT website, and construct the synthetic corpus with \textsc{Base} models that are trained on the parallel data.

Table~\ref{tab:augmentation} lists the results. Both techniques significantly improves the performance of NMT models by exploiting a larger and better phrase table.\footnote{The different sizes of BT and FT phrase tables are due to the different monolingual datasets used for them, the averaged length of which are 24.8 and 28.4, respectively.}
Table~\ref{table:back_transaltion_shared} shows the detailed comparison of the phrase tables. Both augmentation methods {\em induce new knowledge and enhance existing knowledge over the baseline}, and the newly introduced knowledge contribute a lot to the performance improvement.

We further analyze the characteristics of the newly introduced phrase pairs, as illustrated in Figure~\ref{fig:back_translation_shared}. One interesting finding is that the newly introduced phrase pairs are notably longer than the original ones. Besides, the new phrase pairs show less reordered patterns and more monotone patterns, {which may explain the producing of longer phrases. The finding is consistent with previous studies, which show that the BT text is simpler than naturally occurring text \cite{edunov2019evaluation}.}

\subsection{Domain Adaptation}
\label{sec:adaptation}

In the last experiment, we analyze the transferability of the bilingual knowledge by directly applying it to another domain. To this end, we use the IWSLT14 English$\Rightarrow$German data (160,234 sentence pairs) as the target domain (Speech domain), and fine-tune the NMT model trained on the WMT14 dataset (News domain) for several epochs.
We extract the phrase table using the training data of target domain, and the results are shown in  Table~\ref{table:domain-adapt}. Clearly, the fine-tuned NMT model benefits from a larger and better phrase table, by adapting the model to the target domain. The analysis results in Table~\ref{table:shared-domain-adapt} further show that the fine-tuned phrase table improves performance with both more phrases (``Non-Shared'') and better estimation of original phrases (``Shared'').

In addition, we re-extract the phrase from the source domain (WMT data) with the fine-tuned model. 
The phrase table achieves only a BLEU score of 4.77 with 2.6M phrase pairs, while the original model without fine-tune shows a BLEU score of 17.90 with 9.0M phrase pairs. The fine-tune approach increases new knowledge of the target domain while forgets previous-learned knowledge of the source domain.
The results provide an empirical validation of the phenomenon of catastrophic forgetting in domain adaptation~\cite{kirkpatrick2017overcoming}, which inversely demonstrate the reasonableness of our approach.

\begin{table}[t]
\begin{center}
 \begin{tabular}{c|r|r|r|r}
  {\bf Fine}  &  \multicolumn{2}{c|}{\bf NMT} &   \multicolumn{2}{c}{\bf Phrase Table} \\
  \cline{2-5}
  {\bf Tune}  &  \em \# Para.    & \em BLEU  &   \em Size   & \em BLEU  \\
  \hline
  \hline
 \texttimes    &  98M   & 15.78  &  168K  & 16.08 \\
 \checkmark    &  98M   & 31.26  &  316K  & 18.50  \\
 \end{tabular} 
\end{center}
\caption{Results of domain adaptation.}
\label{table:domain-adapt}
\end{table}

\begin{table}[t]
\begin{center}
 \begin{tabular}{c|r|r|r|r}
  {\bf Fine}  &  \multicolumn{2}{c|}{\bf Shared} &   \multicolumn{2}{c}{\bf Non-Shared} \\
  \cline{2-5}
  {\bf Tune}    &  \em Size    & \em BLEU  &   \em Size   & \em BLEU  \\
  \hline
  \hline
  \texttimes    & 0.16M  & 15.95  &  0.01M  &  1.65 \\
  \hline
  \checkmark    &  0.16M & 16.92  &  0.16M  & 6.95  \\
 \end{tabular} 
\end{center}
\caption{Comparison of phrase tables for domain adaptation with or without fine tune.}
\label{table:shared-domain-adapt}
\end{table}

\section{Discussion and Conclusion}

In this work, we propose to assess the bilingual knowledge learned by NMT models with statistic models -- phrase table. The reported results provide a better understanding of NMT models and recent technological advances in learning the essential bilingual lexicons, which also indicate several potential applications:
\begin{itemize}
    \item {\em Error diagnosis} that debugs mistaken predictions by tracing associated phrase pairs~\cite{ding2017visualizing};
    \item {\em Curriculum learning} that dynamically assigns more weights to instances associated with the unlearned knowledge~\cite{platanios2019competence};
    \item {\em Phrase memory} that stores unlearned phrases in NMT to query when generating translations~\cite{wang:2017:aaai,zhang2018prior}.
\end{itemize}

Although the phrase table successfully explains many model behaviors, it cannot explain certain techniques such as enlarging model capacity. The explored bilingual lexicon is only one of the critical knowledge bases in the translation process. In the future, we will investigate more advanced forms of bilingual knowledge~\cite{Liu:2006:ACL,Galley:2010:NAACL}, as well as explore other types of knowledge bases such as grammar and semantics with statistic models (e.g., reordering and language models). This paper is the first step in what we hope will be a long and fruitful journey.

\bibliography{emnlp2020}

\begin{thebibliography}{53}
\expandafter\ifx\csname natexlab\endcsname\relax\def\natexlab#1{#1}\fi

\bibitem[{Alvarez-Melis and Jaakkola(2017)}]{alvarez2017causal}
David Alvarez-Melis and Tommi Jaakkola. 2017.
\newblock A causal framework for explaining the predictions of black-box
  sequence-to-sequence models.
\newblock In \emph{EMNLP}.

\bibitem[{Bahdanau et~al.(2015)Bahdanau, Cho, and Bengio}]{Bahdanau:2015:ICLR}
Dzmitry Bahdanau, Kyunghyun Cho, and Yoshua Bengio. 2015.
\newblock Neural machine translation by jointly learning to align and
  translate.
\newblock In \emph{ICLR}.

\bibitem[{Belinkov et~al.(2017)Belinkov, Durrani, Dalvi, Sajjad, and
  Glass}]{belinkov:2017:ACL}
Yonatan Belinkov, Nadir Durrani, Fahim Dalvi, Hassan Sajjad, and James Glass.
  2017.
\newblock What do neural machine translation models learn about morphology?
\newblock In \emph{ACL}.

\bibitem[{Bengio et~al.(2009)Bengio, Louradour, Collobert, and
  Weston}]{bengio2009curriculum}
Yoshua Bengio, J{\'e}r{\^o}me Louradour, Ronan Collobert, and Jason Weston.
  2009.
\newblock Curriculum learning.
\newblock In \emph{ICML}.

\bibitem[{Brown et~al.(1990)Brown, Cocke, Della~Pietra, Della~Pietra, Jelinek,
  Lafferty, Mercer, and Roossin}]{brown1990statistical}
Peter~F Brown, John Cocke, Stephen~A Della~Pietra, Vincent~J Della~Pietra,
  Frederik Jelinek, John~D Lafferty, Robert~L Mercer, and Paul~S Roossin. 1990.
\newblock A statistical approach to machine translation.
\newblock \emph{Computational linguistics}.

\bibitem[{Brown et~al.(1993)Brown, Pietra, Pietra, and Mercer}]{brown:1993:CL}
Peter~F Brown, Vincent J~Della Pietra, Stephen A~Della Pietra, and Robert~L
  Mercer. 1993.
\newblock The mathematics of statistical machine translation: Parameter
  estimation.
\newblock \emph{Computational linguistics}.

\bibitem[{Chiang(2005)}]{Chiang:2005:ACL}
David Chiang. 2005.
\newblock {A hierarchical phrase-based model for statistical machine
  translation}.
\newblock In \emph{ACL}.

\bibitem[{Ding et~al.(2017)Ding, Liu, Luan, and Sun}]{ding2017visualizing}
Yanzhuo Ding, Yang Liu, Huanbo Luan, and Maosong Sun. 2017.
\newblock Visualizing and understanding neural machine translation.
\newblock In \emph{ACL}.

\bibitem[{Durrani et~al.(2014)Durrani, Haddow, Koehn, and
  Heafield}]{durrani2014edinburgh}
Nadir Durrani, Barry Haddow, Philipp Koehn, and Kenneth Heafield. 2014.
\newblock Edinburgh’s phrase-based machine translation systems for wmt-14.
\newblock In \emph{WMT}.

\bibitem[{Dyer et~al.(2013)Dyer, Chahuneau, and Smith}]{dyer:2013:naacl}
Chris Dyer, Victor Chahuneau, and Noah~A Smith. 2013.
\newblock A simple, fast, and effective reparameterization of ibm model 2.
\newblock In \emph{NAACL}.

\bibitem[{Edunov et~al.(2019)Edunov, Ott, Ranzato, and
  Auli}]{edunov2019evaluation}
Sergey Edunov, Myle Ott, Marc'Aurelio Ranzato, and Michael Auli. 2019.
\newblock On the evaluation of machine translation systems trained with
  back-translation.
\newblock \emph{arXiv preprint arXiv:1908.05204}.

\bibitem[{Galley and Manning(2010)}]{Galley:2010:NAACL}
Michel Galley and Christopher~D. Manning. 2010.
\newblock Accurate non-hierarchical phrase-based translation.
\newblock In \emph{NAACL}.

\bibitem[{Gehring et~al.(2017)Gehring, Auli, Grangier, Yarats, and
  Dauphin}]{gehring17:icml:2017}
Jonas Gehring, Michael Auli, David Grangier, Denis Yarats, and Yann~N. Dauphin.
  2017.
\newblock Convolutional sequence to sequence learning.
\newblock In \emph{ICML}.

\bibitem[{Guo et~al.(2019)Guo, Tan, He, Qin, Xu, and Liu}]{guo2019non}
Junliang Guo, Xu~Tan, Di~He, Tao Qin, Linli Xu, and Tie-Yan Liu. 2019.
\newblock Non-autoregressive neural machine translation with enhanced decoder
  input.
\newblock In \emph{AAAI}.

\bibitem[{Hayes-Roth(1985)}]{hayes1985rule}
Frederick Hayes-Roth. 1985.
\newblock Rule-based systems.
\newblock \emph{Communications of the ACM}, 28(9):921--932.

\bibitem[{He et~al.(2020)He, Gu, Shen, and Ranzato}]{he2019revisiting}
Junxian He, Jiatao Gu, Jiajun Shen, and Marc'Aurelio Ranzato. 2020.
\newblock Revisiting self-training for neural sequence generation.
\newblock In \emph{ICLR}.

\bibitem[{He et~al.(2019)He, Tu, Wang, Wang, Lyu, and Shi}]{He:2019:EMNLP}
Shilin He, Zhaopeng Tu, Xing Wang, Longyue Wang, Michael Lyu, and Shuming Shi.
  2019.
\newblock \href {https://www.aclweb.org/anthology/D19-1088} {Towards
  understanding neural machine translation with word importance}.
\newblock In \emph{EMNLP}.

\bibitem[{Hill et~al.(2017)Hill, Cho, Jean, and Bengio}]{Hill:2017:MT}
Felix Hill, Kyunghyun Cho, S{\'e}bastien Jean, and Yoshua Bengio. 2017.
\newblock The representational geometry of word meanings acquired by neural
  machine translation models.
\newblock \emph{Machine Translation}, 31(1-2):3--18.

\bibitem[{Jain and Wallace(2019)}]{Jain:2019:NAACL}
Sarthak Jain and Byron~C. Wallace. 2019.
\newblock Attention is not explanation.
\newblock In \emph{NAACL}.

\bibitem[{Johnson et~al.(2007)Johnson, Martin, Foster, and
  Kuhn}]{johnson2007improving}
Howard Johnson, Joel Martin, George Foster, and Roland Kuhn. 2007.
\newblock Improving translation quality by discarding most of the phrasetable.
\newblock In \emph{EMNLP}.

\bibitem[{Kirkpatrick et~al.(2017)Kirkpatrick, Pascanu, Rabinowitz, Veness,
  Desjardins, Rusu, Milan, Quan, Ramalho, Grabska-Barwinska
  et~al.}]{kirkpatrick2017overcoming}
James Kirkpatrick, Razvan Pascanu, Neil Rabinowitz, Joel Veness, Guillaume
  Desjardins, Andrei~A Rusu, Kieran Milan, John Quan, Tiago Ramalho, Agnieszka
  Grabska-Barwinska, et~al. 2017.
\newblock Overcoming catastrophic forgetting in neural networks.
\newblock \emph{Proceedings of the national academy of sciences}.

\bibitem[{Kocmi and Bojar(2017)}]{kocmi2017curriculum}
Tom Kocmi and Ond{\v{r}}ej Bojar. 2017.
\newblock Curriculum learning and minibatch bucketing in neural machine
  translation.
\newblock In \emph{RANLP}.

\bibitem[{Koehn(2009)}]{koehn2009statistical}
Philipp Koehn. 2009.
\newblock \emph{Statistical machine translation}.
\newblock Cambridge University Press.

\bibitem[{Koehn et~al.(2007)Koehn, Hoang, Birch, Callison-Burch, Federico,
  Bertoldi, Cowan, Shen, Moran, Zens et~al.}]{koehn2007moses}
Philipp Koehn, Hieu Hoang, Alexandra Birch, Chris Callison-Burch, Marcello
  Federico, Nicola Bertoldi, Brooke Cowan, Wade Shen, Christine Moran, Richard
  Zens, et~al. 2007.
\newblock Moses: Open source toolkit for statistical machine translation.
\newblock In \emph{ACL}.

\bibitem[{Koehn et~al.(2003)Koehn, Och, and Marcu}]{koehn:2003:NAACL}
Philipp Koehn, Franz~Josef Och, and Daniel Marcu. 2003.
\newblock Statistical phrase-based translation.
\newblock In \emph{NAACL}.

\bibitem[{Lample et~al.(2018)Lample, Ott, Conneau, Denoyer, and
  Ranzato}]{lample2018phrase}
Guillaume Lample, Myle Ott, Alexis Conneau, Ludovic Denoyer, and Marc'Aurelio
  Ranzato. 2018.
\newblock Phrase-based \& neural unsupervised machine translation.
\newblock In \emph{EMNLP}.

\bibitem[{Li et~al.(2018)Li, Tu, Yang, Lyu, and Zhang}]{Li:2018:EMNLP}
Jian Li, Zhaopeng Tu, Baosong Yang, Michael~R. Lyu, and Tong Zhang. 2018.
\newblock Multi-head attention with disagreement regularization.
\newblock In \emph{EMNLP}.

\bibitem[{Liu et~al.(2006)Liu, Liu, and Lin}]{Liu:2006:ACL}
Yang Liu, Qun Liu, and Shouxun Lin. 2006.
\newblock {Tree-to-string alignment template for statistical machine
  translation}.
\newblock In \emph{ACL}.

\bibitem[{Lu(2010)}]{lu2010automatic}
Xiaofei Lu. 2010.
\newblock Automatic analysis of syntactic complexity in second language
  writing.
\newblock \emph{International journal of corpus linguistics}.

\bibitem[{Luong and Manning(2015)}]{luong2015stanford}
Minh-Thang Luong and Christopher~D Manning. 2015.
\newblock Stanford neural machine translation systems for spoken language
  domains.
\newblock In \emph{IWSLT}.

\bibitem[{Neubig et~al.(2015)Neubig, Morishita, and Nakamura}]{neubig:2015:wat}
Graham Neubig, Makoto Morishita, and Satoshi Nakamura. 2015.
\newblock Neural reranking improves subjective quality of machine translation:
  Naist at wat2015.
\newblock In \emph{WAT}.

\bibitem[{Och and Ney(2004)}]{Och:2004:CL}
Franz~Josef Och and Hermann Ney. 2004.
\newblock {The alignment template approach to statistical machine translation}.
\newblock \emph{CL}, 30(4):417--449.

\bibitem[{Ott et~al.(2019)Ott, Edunov, Baevski, Fan, Gross, Ng, Grangier, and
  Auli}]{ott:2019:naacl}
Myle Ott, Sergey Edunov, Alexei Baevski, Angela Fan, Sam Gross, Nathan Ng,
  David Grangier, and Michael Auli. 2019.
\newblock Fairseq: A fast, extensible toolkit for sequence modeling.
\newblock \emph{NAACL}.

\bibitem[{Papineni et~al.(2002)Papineni, Roukos, Ward, and
  Zhu}]{papineni2002bleu}
Kishore Papineni, Salim Roukos, Todd Ward, and Wei-Jing Zhu. 2002.
\newblock Bleu: a method for automatic evaluation of machine translation.
\newblock In \emph{ACL}.

\bibitem[{Platanios et~al.(2019)Platanios, Stretcu, Neubig, Poczos, and
  Mitchell}]{platanios2019competence}
Emmanouil~Antonios Platanios, Otilia Stretcu, Graham Neubig, Barnabas Poczos,
  and Tom Mitchell. 2019.
\newblock Competence-based curriculum learning for neural machine translation.
\newblock In \emph{NAACL}.

\bibitem[{Rahaman et~al.(2019)Rahaman, Baratin, Arpit, Draxler, Lin, Hamprecht,
  Bengio, and Courville}]{Rohaman:2019:ICML}
Nasim Rahaman, Aristide Baratin, Devansh Arpit, Felix Draxler, Min Lin, Fred
  Hamprecht, Yoshua Bengio, and Aaron Courville. 2019.
\newblock On the spectral bias of neural networks.
\newblock In \emph{ICML}.

\bibitem[{Sato(1992)}]{sato1992example}
Satoshi Sato. 1992.
\newblock Example-based machine translation.

\bibitem[{Sennrich et~al.(2016)Sennrich, Haddow, and Birch}]{sennrich:2016:acl}
Rico Sennrich, Barry Haddow, and Alexandra Birch. 2016.
\newblock Neural machine translation of rare words with subword units.
\newblock In \emph{ACL}.

\bibitem[{Shi et~al.(2016)Shi, Padhi, and Knight}]{shi:2016:EMNLP}
Xing Shi, Inkit Padhi, and Kevin Knight. 2016.
\newblock Does string-based neural {MT} learn source syntax?
\newblock In \emph{EMNLP}.

\bibitem[{Sutskever et~al.(2014)Sutskever, Vinyals, and
  Le}]{Sutskever:2014:NIPS}
Ilya Sutskever, Oriol Vinyals, and Quoc~V Le. 2014.
\newblock {Sequence to sequence learning with neural networks}.
\newblock In \emph{NIPS}.

\bibitem[{Tillmann(2004)}]{tillmann:2004:naacl}
Christoph Tillmann. 2004.
\newblock A unigram orientation model for statistical machine translation.
\newblock In \emph{HLT-NAACL}.

\bibitem[{Toneva et~al.(2019)Toneva, Sordoni, Combes, Trischler, Bengio, and
  Gordon}]{toneva:2018:empirical}
Mariya Toneva, Alessandro Sordoni, Remi Tachet~des Combes, Adam Trischler,
  Yoshua Bengio, and Geoffrey~J Gordon. 2019.
\newblock An empirical study of example forgetting during deep neural network
  learning.

\bibitem[{Tu et~al.(2016)Tu, Lu, Liu, Liu, and Li}]{Tu_2016}
Zhaopeng Tu, Zhengdong Lu, Yang Liu, Xiaohua Liu, and Hang Li. 2016.
\newblock Modeling coverage for neural machine translation.
\newblock In \emph{ACL}.

\bibitem[{Vaswani et~al.(2017)Vaswani, Shazeer, Parmar, Uszkoreit, Jones,
  Gomez, Kaiser, and Polosukhin}]{Vaswani:2017:NIPS}
Ashish Vaswani, Noam Shazeer, Niki Parmar, Jakob Uszkoreit, Llion Jones,
  Aidan~N Gomez, {\L}ukasz Kaiser, and Illia Polosukhin. 2017.
\newblock {Attention is All You Need}.
\newblock In \emph{NIPS}.

\bibitem[{Voita et~al.(2019{\natexlab{a}})Voita, Sennrich, and
  Titov}]{voita:2019:EMNLP}
Elena Voita, Rico Sennrich, and Ivan Titov. 2019{\natexlab{a}}.
\newblock The bottom-up evolution of representations in the transformer: A
  study with machine translation and language modeling objectives.
\newblock In \emph{EMNLP}.

\bibitem[{Voita et~al.(2019{\natexlab{b}})Voita, Talbot, Moiseev, Sennrich, and
  Titov}]{Voita:2019:ACL}
Elena Voita, David Talbot, Fedor Moiseev, Rico Sennrich, and Ivan Titov.
  2019{\natexlab{b}}.
\newblock \href {https://www.aclweb.org/anthology/P19-1580} {Analyzing
  multi-head self-attention: Specialized heads do the heavy lifting, the rest
  can be pruned}.
\newblock In \emph{ACL}.

\bibitem[{Wang et~al.(2017)Wang, Lu, Tu, Li, Xiong, and Zhang}]{wang:2017:aaai}
Xing Wang, Zhengdong Lu, Zhaopeng Tu, Hang Li, Deyi Xiong, and Min Zhang. 2017.
\newblock Neural machine translation advised by statistical machine
  translation.
\newblock In \emph{AAAI}.

\bibitem[{{Wang} et~al.(2018){Wang}, {Tu}, and {Zhang}}]{Wang:2018:TASLP}
Xing {Wang}, Zhaopeng {Tu}, and Min {Zhang}. 2018.
\newblock Incorporating statistical machine translation word knowledge into
  neural machine translation.
\newblock \emph{IEEE/ACM Transactions on Audio, Speech, and Language
  Processing}, 26(12):2255--2266.

\bibitem[{Wiegreffe and Pinter(2019)}]{Wiegreffe:2019:EMNLP}
Sarah Wiegreffe and Yuval Pinter. 2019.
\newblock \href {https://www.aclweb.org/anthology/D19-1002} {Attention is not
  not explanation}.
\newblock In \emph{EMNLP}.

\bibitem[{Yang et~al.(2019)Yang, Wang, Wong, Chao, and Tu}]{Yang:2019:ACL}
Baosong Yang, Longyue Wang, Derek~F. Wong, Lidia~S. Chao, and Zhaopeng Tu.
  2019.
\newblock Convolutional self-attention networks.
\newblock In \emph{ACL}.

\bibitem[{Zhang et~al.(2017)Zhang, Liu, Luan, Xu, and Sun}]{zhang2018prior}
Jiacheng Zhang, Yang Liu, Huanbo Luan, Jingfang Xu, and Maosong Sun. 2017.
\newblock Prior knowledge integration for neural machine translation using
  posterior regularization.
\newblock In \emph{ACL}.

\bibitem[{Zhang and Zong(2016)}]{zhang2016exploiting}
Jiajun Zhang and Chengqing Zong. 2016.
\newblock Exploiting source-side monolingual data in neural machine
  translation.
\newblock In \emph{EMNLP}.

\bibitem[{Zhao et~al.(2018)Zhao, Wang, Zhang, and Zong}]{zhao2018phrase}
Yang Zhao, Yining Wang, Jiajun Zhang, and Chengqing Zong. 2018.
\newblock Phrase table as recommendation memory for neural machine translation.
\newblock In \emph{IJCAI}.

\end{thebibliography}
\bibliographystyle{acl_natbib}

\end{document}